\documentclass[sigconf]{acmart}
\AtBeginDocument{%
  \providecommand\BibTeX{{%
    \normalfont B\kern-0.5em{\scshape i\kern-0.25em b}\kern-0.8em\TeX}}}

\copyrightyear{2022}
\acmYear{2022}
\setcopyright{acmcopyright}\acmConference[MM '22]{Proceedings of the 30th ACM
International Conference on Multimedia}{October 10--14, 2022}{Lisboa, Portugal}
\acmBooktitle{Proceedings of the 30th ACM International Conference on Multimedia
(MM '22), October 10--14, 2022, Lisboa, Portugal}
\acmPrice{15.00}
\acmDOI{10.1145/3503161.3548214}
\acmISBN{978-1-4503-9203-7/22/10}

\acmSubmissionID{1899}

\usepackage{adjustbox}
\usepackage{threeparttable}
\usepackage{tablefootnote}
\usepackage{graphicx}
\usepackage{amsmath}
\usepackage{amssymb}
\usepackage{multirow}
\usepackage{algorithm}
\usepackage{algorithmic}

\begin{document}

\title{Marior: Margin Removal and Iterative Content Rectification
for Document Dewarping in the Wild}

\author{Jiaxin Zhang}
\author{Canjie Luo}
\affiliation{%
  \institution{South China University of Technology}
  \country{}}
\email{msjxzhang@mail.scut.edu.cn}
\email{canjie.luo@gmail.com}

\author{Lianwen Jin}
\authornote{Corresponding author.}
\affiliation{%
  \institution{South China University of Technology}
  \institution{Peng Cheng Laboratory}
  \institution{Pazhou Laboratory (Huangpu)}
  \country{}}
\email{eelwjin@scut.edu.cn}

\author{Fengjun Guo}
\author{Kai Ding}
\authornotemark[1]
\affiliation{%
  \institution{IntSig Information Co. Ltd}
  \country{}}
\email{fengjun_guo@intsig.net}  
\email{danny_ding@intsig.net}

\def\authors{Jiaxin Zhang, Canjie Luo, Lianwen Jin, Fengjun Guo, Kai Ding}

\renewcommand{\shortauthors}{Jiaxin Zhang, et al.}

\begin{abstract}
  Camera-captured document images usually suffer from perspective and geometric deformations. It is of great value to rectify them when considering poor visual aesthetics and the deteriorated performance of OCR systems. Recent learning-based methods intensively focus on the accurately cropped document image. However, this might not be sufficient for overcoming practical challenges, including document images either with large marginal regions or without margins. Due to this impracticality, users struggle to crop documents precisely when they encounter large marginal regions. Simultaneously, dewarping images without margins is still an insurmountable problem. To the best of our knowledge, there is still no complete and effective pipeline for rectifying document images in the wild. To address this issue, we propose a novel approach called Marior (\textbf{Ma}rgin \textbf{R}emoval and \textbf{I}terative C\textbf{o}ntent \textbf{R}ectification). Marior follows a progressive strategy to iteratively improve the dewarping quality and readability in a coarse-to-fine manner. Specifically, we divide the pipeline into two modules: margin removal module (MRM) and iterative content rectification module (ICRM). First, we predict the segmentation mask of the input image to remove the margin, thereby obtaining a preliminary result. Then we refine the image further by producing dense displacement flows to achieve content-aware rectification. We determine the number of refinement iterations adaptively. Experiments demonstrate the state-of-the-art performance of our method on public benchmarks. The resources are available at \url{https://github.com/ZZZHANG-jx/Marior} for further comparison.
\end{abstract}




\begin{teaserfigure}
  \includegraphics[scale=0.31]{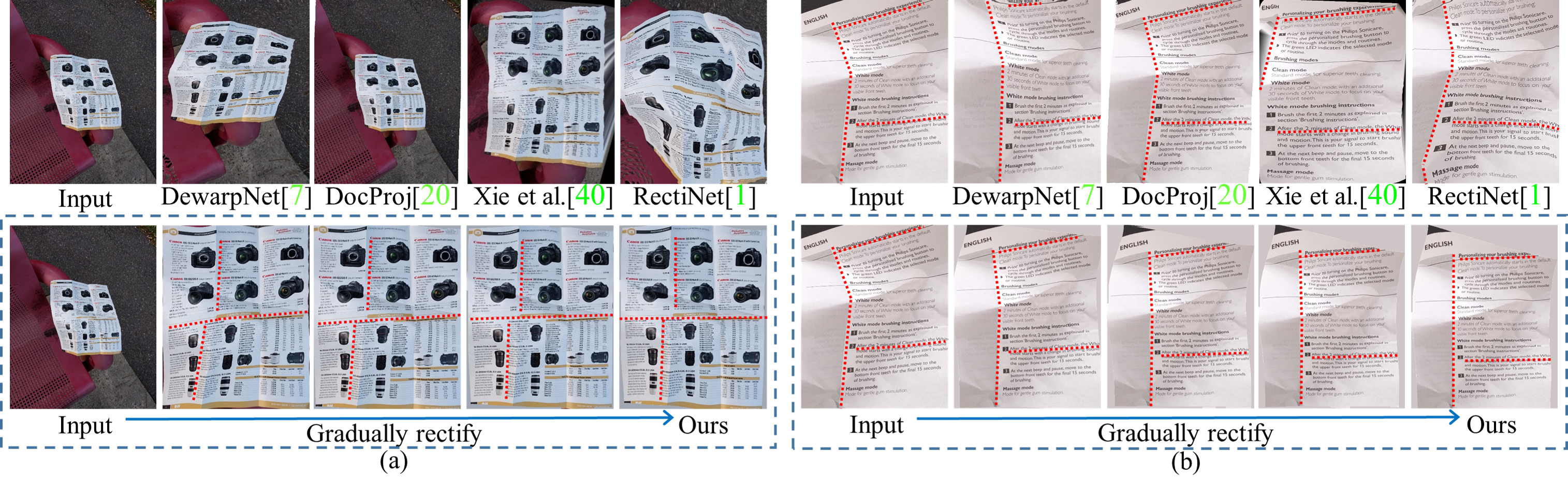}
  \centering
  \vspace{-3mm}
  \caption{Cases that have not been sufficiently studied by existing deep learning-based methods. (a) Document image with large marginal regions. (b) Document image without marginal regions. Marginal region refers to the area composed of pixels that do not belong to the document of interest. We highlight the deformation using red dotted lines.}
  \label{fig:first_example}
\end{teaserfigure}

\maketitle

\section{Introduction}
\label{sec:intro}

Powered by advanced built-in cameras in mobile devices, digitizing ubiquitous documents in daily life has become convenient for people. However, because of the inappropriate angle and position of the camera, the captured document images usually contain perspective deformation. Besides, the document itself may also geometrically deform because of curving, folding, or creasing. These types of deformation lead to the deteriorated performance of optical character recognition (OCR) systems and poor readability for readers.

Recent deep learning-based dewarping methods~\cite{ma2018docunet,das2019dewarpnet,xie2020dewarping,augn,das2021end,garai2021theoretical,xie2021document,ramanna2019document,garai2021dewarping,feng2021docscanner,feng2021doctr} have made great progress regarding robustness to a variety of document layouts. However, almost all of them only focus on accurately cropped document images, and ignore cases with large marginal regions or without marginal regions, which are shown in Fig.~\ref{fig:first_example} (a) and (b), respectively. In this study, marginal region refers to the area composed of pixels that do not belong to the document of interest. To address this issue, one can take all these situations into consideration during training, but we found the results unsatisfactory (refer to supplementary material). We argue that this is attributed to the extra implicit learning to identify the foreground document and remove marginal region, which is also observed in \cite{feng2021docscanner}. Another way is to implement en existing object detection algorithm ahead of dewarping to avoid the need for manually cropping. However, dewarping document images without margins is still an unresolved problem. Accordingly, there is still no complete and effective pipeline for handling all cases in the wild. 

Therefore, we propose Marior (\textbf{Ma}rgin \textbf{R}emoval and \textbf{I}terative C\textbf{o}ntent \textbf{R}ectification) to tackle this problem, which consists of two cascaded modules: margin removal module (MRM) and iterative content rectification module (ICRM). Marrior decouples the margin removal and document rectification process. Specifically, in the MRM, we first feed the source distorted image into our mask prediction network, which predicts the corresponding document segmentation mask. Then we propose a novel mask-based dewarper (MBD) to remove the margin based on this mask and obtain a preliminary dewarped result. For images without marginal regions and no complete document edges, as shown in Fig.~\ref{fig:first_example} (b), we propose filtering them out and skipping the margin removal process by using an intersection-over-union (IoU)-based method, which is inspired by the observation that these images usually result in noisy masks.

Thereafter, we feed the margin-removed output from the MRM into the ICRM for further refinement. It predicts a dense displacement flow that assigns a two{-}dimensional (2D) offset vector to each pixel in the input image. After rectification according to this flow, we obtain a dewarped output image. Because the margin-removed image focuses more on content (\textit{e.g.}, text lines and figures), the ICRM should be content-aware. Therefore, we further design a new content-aware loss to implicitly guide the ICRM to focus more on informative regions, such as text lines and figures, than the uniform document background. This design is based on an intuition that the latter contains fewer deformation clues and slight deviation for it on dewarped result is visually negligible. In addition, we have found that iterative implementation of ICRM can improve rectification performance. To this end, we propose an adaptive approach to determine the number of iterations to make the proposed iterative ICRM process more intelligent and efficient.

To summarize, our contributions are as follows:

\begin{itemize}
  \item We propose a new method called Marior to handle document images that have various margin situations, which are ignored by existing learning-based methods. 
  \item We propose a new mask-based dewarper in our margin removal module (MRM) that coarsely dewarps the document image based on the predicted segmentation mask. Then an iterative content rectification module (ICRM) is proposed to further refine the image by predicting a dense displacement flow. 
  \item We design a new content-aware loss to implicitly guide the flow prediction network to focus more on informative regions. We also propose an adaptive iteration strategy to improve performance.
  \item Extensive experiments demonstrate that the proposed Marior achieves state-of-the-art performance on two widely used public benchmarks. Moreover, this approach achieves significant success for tackling cases that have diverse margins.
\end{itemize}

\section{Related work}

\subsection{Traditional hand-crafted methods}
Document image dewarping, or geometric rectification, has been widely studied in the literature. Most traditional hand-crafted methods attempted to reconstruct the three-dimensional (3D) shape of document images. The most direct approach is to use additional equipment, such as a depth sensor~\cite{2008An}, visible light projector-camera system~\cite{2001Document}, and structured laser light sources~\cite{meng2014active}. This type of method can achieve significant performance in specific scenarios, but its application is very limited. Methods in~\cite{koo2009composition,ld2017,tsoi2007multi} used multiple images from different views to reconstruct the 3D shape rather than relying on additional equipment. However, they are not sufficiently practical because multiple images are not always available in many situations. To reconstruct the 3D shape from a single image without additional equipment, researchers used low-level visual cues in document images, such as illumination effects~\cite{tan2005restoring,zhang2009unified}, text lines~\cite{tian2011rectification,liu2015restoring}, and boundaries~\cite{brown2006geometric,cao2003cylindrical}. Such low-level visual cues were also widely used in some 2D image processing-based methods~\cite{zhang2008arbitrary,stamatopoulos2010goal,gatos2007segmentation,guo2019fast} that did not need to reconstruct the 3D shape. Methods that relied on these low-level visual cues made strong assumptions about the document layout or deformation type, which were not sufficiently robust for real-world scenes.

\begin{figure*}[t]
  \centering
  \includegraphics[scale=0.39]{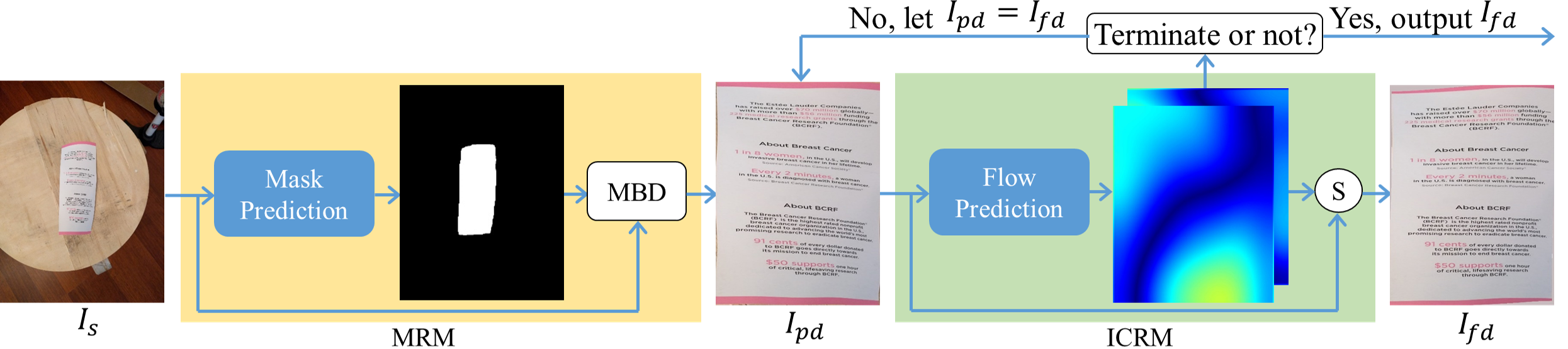}
  \caption{Overall pipeline of our proposed Marior. $I_{s}$, $I_{pd}$, and $I_{fd}$ denote the source distorted image, preliminary dewarped image, and final dewarped image respectively. Marior takes $I_{s}$ as input and gradually rectifies it to output $I_{fd}$. MBD is a mask-based dewarper and $S$ denotes the sampling procedure.}
  \label{fig:pipeline}
\end{figure*}

\subsection{Learning-based methods}
Learning-based methods are becoming increasingly popular because of the availability of large datasets~\cite{das2019dewarpnet,bandyopadhyay2021gated,ma2018docunet}.
Das \textit{et al.}~\cite{das2017common} used convolutional neural networks (CNNs) in this task for the first time, where they only adopted CNNs to detect creases to help subsequent 2D image processing-based methods. More studies have emerged since the proposal of two synthetic datasets~\cite{ma2018docunet,das2019dewarpnet}. These two datasets contain pixel-wise unwarping map annotation that can be directly used to sample the dewarped image from the source distorted image. Ma \textit{et al.}~\cite{ma2018docunet} used a stacked U-Net to obtain a refined forward map. Das \textit{et al.}~\cite{das2019dewarpnet} used several pixel-wise annotations from their Doc3D dataset to train CNNs to predict the 3D coordinates, backward map, and shading map, which were used to dewarp the image and adjust the illumination. Liu~\cite{augn} proposed a multi-resolution method and a generative adversarial network (GAN) framework to make the output more visually pleasing. Li \textit{et al.}~\cite{li2019document} proposed a patch-based method to split the input image into small patches both during training and inference, thus reducing the learning difficulty and focusing more on details. However, the patch stitching process was very time-consuming. 
\begin{figure}[!t]
  \centering
  \includegraphics[scale=0.16]{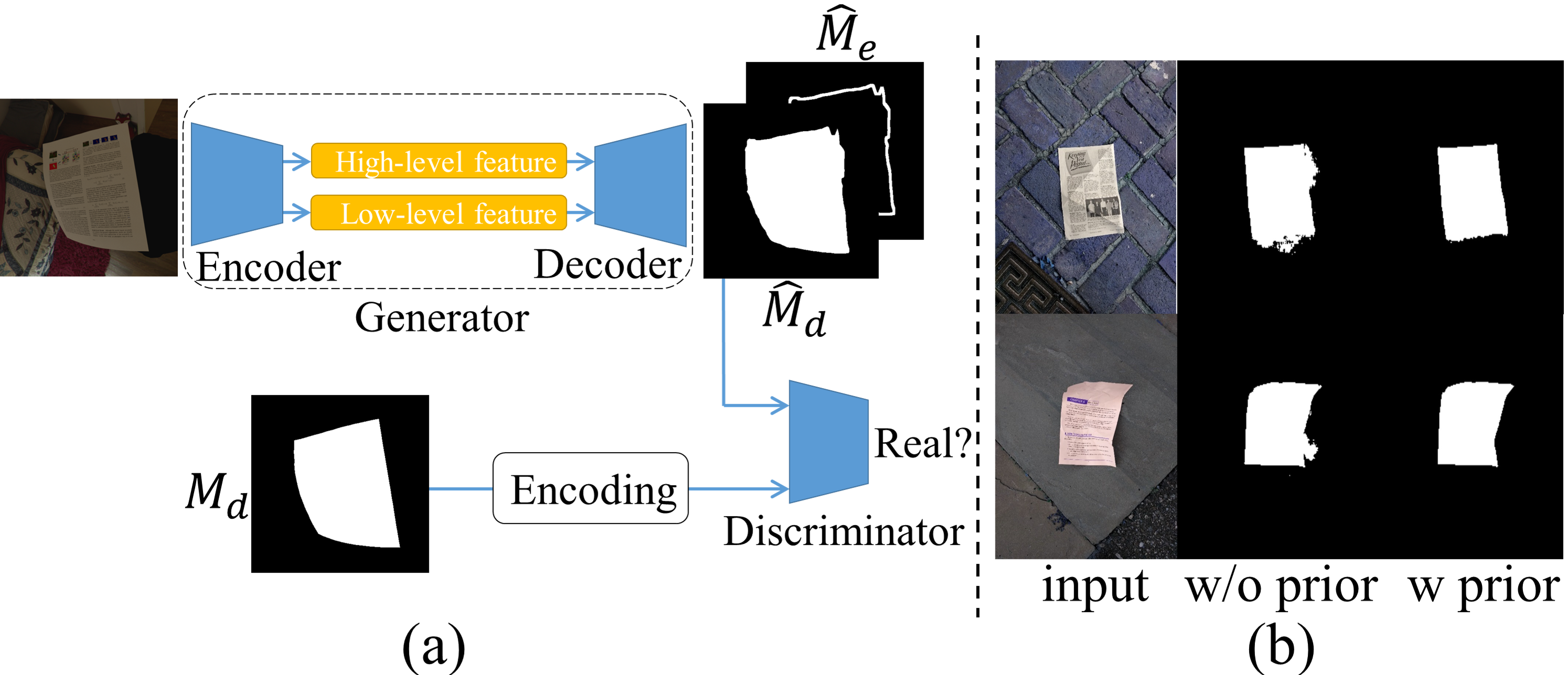}
  \caption{(a) Architecture of our mask prediction network. (b) Effectiveness by introducing prior knowledge. The prior knowledge brings clean and reasonable masks.}
  \label{fig:MRM}
  \vspace{-3mm}
\end{figure}
Das \textit{et al.}~\cite{das2021end} proposed an end-to-end trainable patch-based method to alleviate that problem and improve performance. Amir \textit{et al.}~\cite{markovitz2020can} explored the document content to guide the optimization of the neural network. Feng \textit{et al.}~\cite{feng2021doctr} used the Transformer~\cite{vaswani2017attention} as network architecture and achieved superior performance. However, these methods can only handle accurately cropped document images. There is still no complete and effective pipeline that is sufficiently robust to dewarp document images in the wild. Even for accurately cropped cases, the dewarping performance of these methods are still subpar for practical applications.

\section{Methodology}
As shown in Fig.~\ref{fig:pipeline}, Marior contains two cascaded MBD and ICRM modules which progressively rectify the distorted source image $I_{S}$ and output the final dewarped image $I_{fd}$. In the MRM, we first remove the margin based on the predicted mask and obtain a preliminary dewarped result $I_{pd}$. This mask-based dewarping process is achieved using a novel MBD. Then the ICRM takes as input $I_{pd}$ and predicts a dense displacement flow that has the same resolution as $I_{pd}$. This 2D flow assigns the distance that each pixel in $I_{pd}$ should be shifted to obtain $I_{fd}$. We sample $I_{fd}$ from $I_{pd}$ based on this flow. To gain better rectification performance, we implement the ICRM iteratively. We propose an adaptive method to determine the number of iterations.

\subsection{Margin removal module (MRM)}
\textbf{Mask prediction}. To remove the margin from a given image, we first localize document regions. We consider the localization as a semantic segmentation task, which aims to produce a mask that precisely represents document regions. The architecture of our mask prediction network is shown in Fig.~\ref{fig:MRM} (a). We directly adopt the encoder and decoder from DeepLabv3+~\cite{chen2018encoder}. In addition to the document mask, we design a head to produce an edge mask for auxiliary training. Furthermore, we observe that the document mask has a unique and relatively fixed pattern, such as relatively straight edges, one large connected area, and a shape close to a quadrilateral. We impose this prior knowledge in the MRM using a GAN framework as shown in Fig.~\ref{fig:MRM} (a). We find this can effectively reduce noise on the produced mask, as shown in Fig.~\ref{fig:MRM} (b). The objective is defined as
\begin{equation}\small
  \min_{MRM} \mathcal{L}_{MRM} = \min_{MRM} (\lambda \mathcal{L}_{\text {prior }}+\mathcal{L}_{\text {mask }}+\mathcal{L}_{\text {edge }}),
\end{equation} 
\begin{equation}\small
  \max_{D_{net}} \mathcal{L}_{D_{net}}= \max_{D_{net}} \mathcal{L}_{\text {prior }}.
\end{equation}
$\mathcal{L}_{\text {mask}}$ and $\mathcal{L}_{\text {edge}}$ are standard binary cross-entropy losses:
\begin{equation}\small
\begin{aligned}
  \mathcal{L}_{\text {mask }}=-\frac{1}{N} \sum_{i}^{N}&[m_{d_{i}} \cdot \log \left(\widehat{m}_{d_{i}}\right) \\
  &+\left(1-m_{d_{i}}\right) \cdot \log \left(1-\widehat{m}_{d_{i}}\right)],
\end{aligned}
\end{equation}
\begin{equation}\small
\begin{aligned}
  \mathcal{L}_{e d g e}=-\frac{1}{N} \sum_{i}^{N}&[m_{e_{i}} \cdot \log \left(\widehat{m}_{e_{i}}\right)\\
  &+\left(1-m_{e_{i}}\right) \cdot \log \left(1-\widehat{m}_{e_{i}}\right)],
\end{aligned}
\end{equation}
where $\widehat{m}_{d_{i}}$ and $\widehat{m}_{e_{i}}$ denote the predicted classification for the $i$-th element in document mask $\widehat{M}_{d}$ and edge mask $\widehat{M}_{e}$, and $m_{d}$ and $m_{e}$ are their corresponding ground truths, respectively. $N$ is the number of elements in $\widehat{M}_{d}$. $\mathcal{L}_{\text {prior}}$ is a standard objective in GAN framework that guides the distribution of $\widehat{M}_{d}$ closer to that of ground-truth mask $M_{d}$, and $\lambda$ is the weight for this term:
\begin{equation}\small
  \mathcal{L}_{\text {prior }}=\mathbb{E}\left[\log D_{net}\left(M_{d}^{\prime}\right)\right]+\mathbb{E}\left[\log \left(1-D_{net}\left(\widehat{M}_{d}\right)\right)\right].
\end{equation}
Inspired by~\cite{luc2016semantic}, we replace the one-hot coding of $M_{d}$ with $M_{d}^{\prime}$. We denote the $i$-th element in $M_{d}^{\prime}$ by 
\begin{equation}\small
  m_{d_{i}}^{\prime}=\left\{\begin{array}{l}
  \max \left(0.9, \widehat{m}_{d_{i}}\right), m_{d_{i}}=1 \\
  \min \left(0.1, \widehat{m}_{d_{i}}\right), m_{d_{i}}=0
  \end{array} \quad i=1,2, \ldots, N\right..
\end{equation}
\begin{figure}[!t]
  \centering
  \includegraphics[scale=0.28]{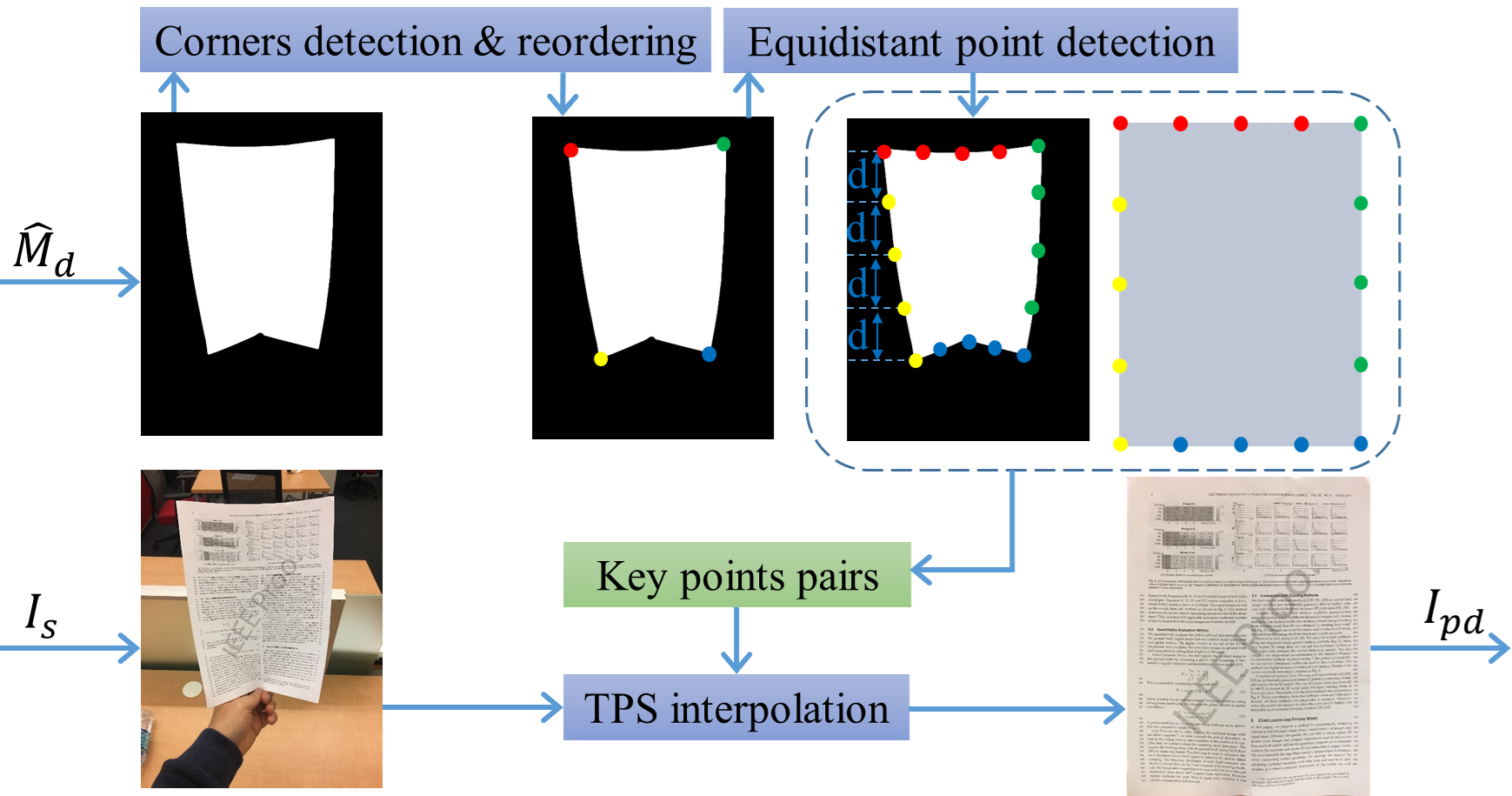}{}
  \caption{Proposed mask-based dewarper (MBD), which takes as input $I_{s}$ and $\widehat{M}_{d}$, and outputs $I_{pd}$.}
  \label{fig:mbd}
  \vspace{-5mm}
\end{figure}
This aims to reduce the distribution gap between the one-hot coding positive sample and the generated negative sample under when optimizing the discriminator. It is worth noting that this mask prediction model can also be other alternative segmentation models, which just need to be able to provide the segmentation mask of document regions.

\textbf{Mask-based dewarper (MBD)}. After obtaining the document mask, we propose a new MBD to remove the margin and perform preliminary dewarping, as shown in Fig.~\ref{fig:mbd}. Specifically, based on the predicted mask, we first detect the four corners using the Douglas{-}Peucker algorithm~\cite{saalfeld1999topologically} and then determine the order (left top, right top, right bottom, and left bottom) based on their relative positions. Then we can determine equidistant points on each edge (in our experiment, in addition to the four corners, we use three equidistant points on each edge). We match these control points to the corresponding positions of a rectangle. Then we use these key point pairs to perform thin plate spline (TPS)~\cite{warps1989thin} interpolation on $I_{s}$, thereby removing the margin and obtaining $I_{pd}$. Notably, for document images without marginal regions, which do not have complete edges, as shown in Fig.~\ref{fig:first_example} (b), we skip TPS interpolation and consider the original $I_{s}$ as the output of the MBD. We filter these images out by calculating the IoU between $\widehat{M}_{d}$ and the mask derived from all detected control points, and setting a threshold for this IoU. This is inspired by the observation that document images without complete edges usually lead to noisy $\widehat{M}_{d}$, thus resulting in relatively low IoU.

\subsection{Iterative content rectification module (ICRM)}
The results from preliminary dewarping using the MRM are not perfect. The reasons are twofold. The first reason is that the selection of equidistant points on each edge does not consider depth information; hence, this equidistant division is inconsistent with that performed on physical paper. The second reason is that, sometimes, the predicted mask is not sufficiently accurate when it encounters unclear edges or very complex margins. Additionally, document images without marginal regions skip preliminary dewarping, thereby still being untouched.

\begin{figure}[!t]
  \centering
  \includegraphics[scale=0.123]{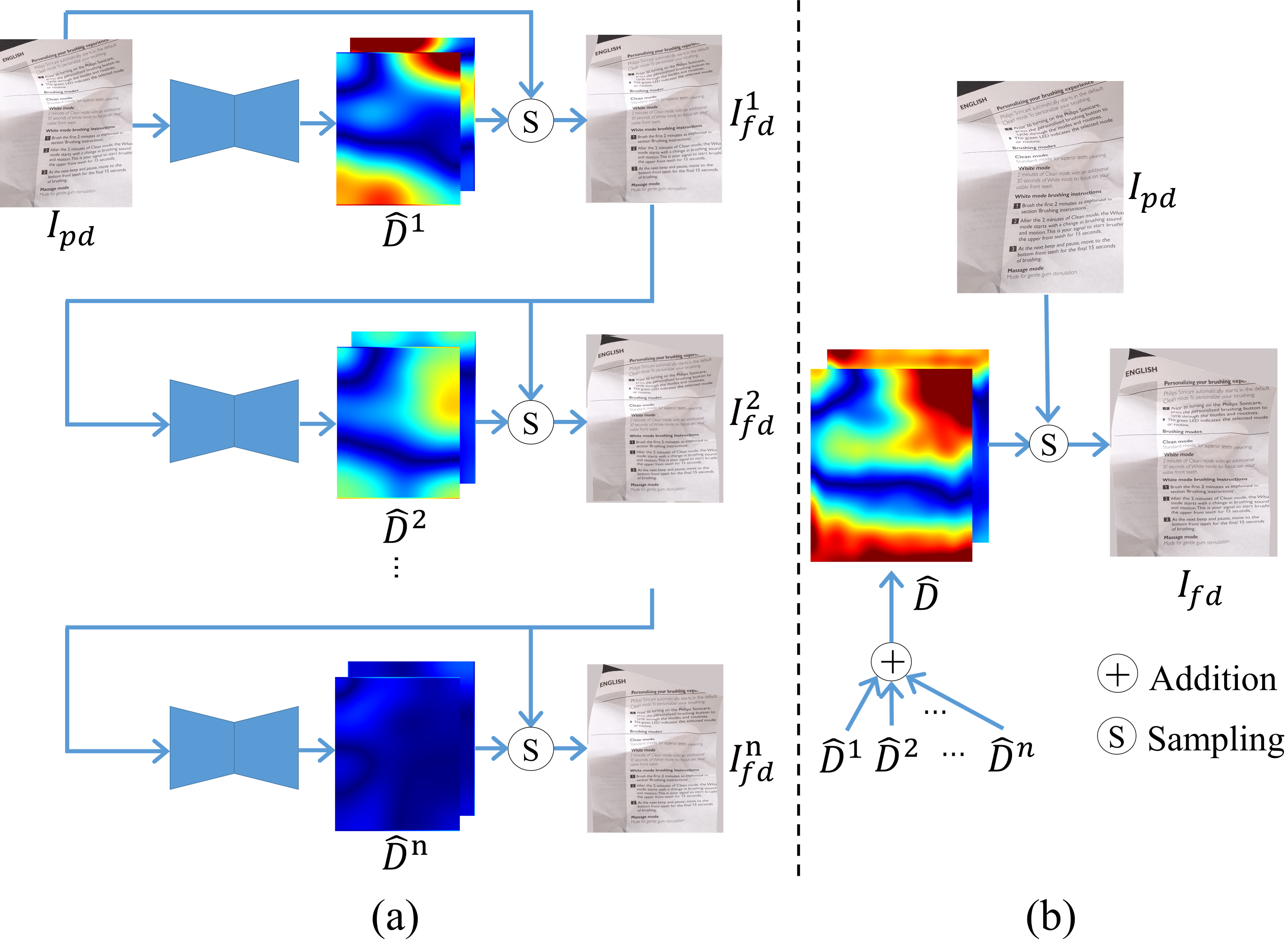}{}
  \caption{(a) Iterative rectification scheme of the proposed ICRM. (b) Sampling procedure after the termination of iteration . The intensity of the heat map indicates the absolute value of the displacement flow, that is, distance should be shifted. Red regions correspond to a large value and blue regions correspond to a small value.}
  \label{fig:icrm}
  \vspace{-2mm}
\end{figure}

To further rectify $I_{pd}$, we propose the ICRM, which takes $I_{pd}$ as input and produces a dense displacement flow $\widehat{D}$. We adopt the commonly used encoder-decoder with skip connections as our displacement flow prediction network. We adopt the attention strategy~\cite{woo2018cbam} in bottleneck and dilated convolution~\cite{yu2015multi} to enlarge receptive filed to capture global information. As mentioned previously, the rectification of informative regions, such as text lines and figures, is intuitively more crucial than that of the uniform document background. We use the document content mask $M_{c}$ to design our content-aware loss $\mathcal{L}_{c}$, which implicitly guides the network to focus more on informative regions. We also adopt shift invariant loss $\mathcal{L}_{s}$ as in ~\cite{ma2018docunet}. The final training loss of ICRM is expressed as
\begin{equation}\small
  \label{eq:icrm_loss1}
  \mathcal{L}_{ICRM}=\mathcal{L}_{c}+\alpha \mathcal{L}_{s},
\end{equation}
\begin{equation}\small
  \label{eq:icrm_loss}
  \mathcal{L}_{c}=\frac{1}{N} \sum_{i}^{N}\left(\left\|d_{i}-\hat{d}_{i}\right\|_{2}+\beta \cdot m_{c_{i}} \cdot\left\|d_{i}-\hat{d}_{i}\right\|_{2}\right),
\end{equation}
\begin{equation}\small
\mathcal{L}_{s}=\frac{1}{2 N^{2}} \sum_{i, j}\left(\left(d_{i}-d_{j}\right)-\left(\hat{d}_{i}-\hat{d}_{j}\right)\right)^{2},
\end{equation}
where $\hat{d}_{i}$, $d_{i}$, and $m_{c_{i}}$ denote the $i$-th element in predicted displacement flow $\widehat{D}$, ground-truth displacement flow $D$, and document content mask $M_{c}$, respectively. $\alpha$ and $\beta$ are constant weights.

Because we complete margin removal in the MRM, the ICRM is supposed to focus on content rectification without extra implicit learning to identity the foreground document and remove marginal region. The separation of margin removal also makes the ICRM capable of adopting an iterative scheme to rectify the document step by step, which we find can improve rectification performance. If margin removal is not decoupled, the network might learn to rectify the document based on document edges and tend to find them in every iteration, even if they do not exist, which will result in problematic output. Our iterative scheme is shown in Fig.~\ref{fig:icrm} (a). In the beginning, we feed $I_{pd}$ into the displacement flow prediction network and obtain the first displacement flow $\widehat{D}^{1}$, which we can then use to sample $I_{f d}^{1}$ from $I_{pd}$:

\begin{equation}\small
  I_{f d}^{1}=\mathrm{S}\left(I_{p d}, \widehat{D}^{1}\right), \text{where } \widehat{D}^{1}=\mathrm{CNN}\left(I_{p d}\right),
\end{equation}
where $S$ denotes the sampling procedure. As shown in Fig.~\ref{fig:icrm} (a), deformation remains in $I_{f d}^{1}$. We apply iterative scheme to further refine the rectification result according to the following equation:

\begin{equation}\small
  I_{f d}^{n}=\mathrm{S}\left(I_{p d}, \widehat{D}^{n}\right), \text{where } \widehat{D}^{n}=\mathrm{CNN}\left(I_{f d}^{n-1}\right).
\end{equation}

After several iterations, $I_{f d}^{n}$ already achieves satisfactory rectification performance. The response in $\widehat{D}^{n}$ has significantly decreased because of the relatively flat input $I_{f d}^{n-1}$. More iterations consume more time and even introduce new distortions. Hence, the iteration procedure should terminate at the appropriate time. We propose an adaptive method to determine this time based on the statistics of the flow, as shown in Algorithm ~\ref{alg:alg1}.

\begin{minipage}{7cm}
\begin{algorithm}[H]
	\caption{Decide whether to terminate iteration.} 
	\label{alg3} 
	\begin{algorithmic}
		\IF{$\operatorname{var}\left(\widehat{D}^{n}\right) > \operatorname{var}\left(\widehat{D}^{n-1}\right)$} 
		\STATE terminate
		\ELSIF{$\operatorname{var}\left(\widehat{D}^{n}\right) \leq \tau$}
		\STATE terminate
		\ELSE 
		\STATE continue
		\ENDIF 
	\end{algorithmic}
	\label{alg:alg1}
\end{algorithm}
\vspace{0.5mm}
\end{minipage}

\noindent Here $\operatorname{var}\left(\widehat{D}^{n}\right)$ is the variance of $\widehat{D}^{n}$ and $\tau$ is a pre-defined constant value that serves as the threshold. After the termination of the iterative procedure, we obtain the final displacement flow $\widehat{D}$ by summing all previous $\widehat{D}^{i}(\mathrm{i}=1,2 \ldots, \mathrm{n})$ and obtain the final dewarped result $I_{fd}$ based on $\widehat{D}$:
\begin{equation}\small
  I_{f d}=\mathrm{S}\left(I_{p d}, \widehat{D}\right), \text{where } \widehat{D}=\sum_{i=1}^{n} \widehat{D}^{i}.
\end{equation}

\section{Experiments}
\subsection{Dataset}
We train both networks in the MRM and ICRM on the Doc3D~\cite{das2019dewarpnet} dataset, which consists of 100k rich-annotated samples. We split the dataset into 90k training data and 10k validation data. During the training of mask prediction, we randomly replace the margin with texture images from~\cite{cimpoi2014describing} as data augmentation. In addition to commonly used random cropping and scaling, we also adopt random erasing~\cite{zhong2020random}. The training data (both the source distorted input image and ground-truth displacement flow) for the ICRM are first preprocessed by our proposed MBD. We consider the binarization result derived from the albedo map (provided in Doc3D~\cite{das2019dewarpnet}) as $M_{c}$ in Eq.~\ref{eq:icrm_loss}.

\footnotetext[1]{\scriptsize \url{https://github.com/tesseract-ocr/tesseract}}

\subsection{Implementation details}
We implement our model in the PyTorch framework~\cite{pytorch} and train it on a single NVIDIA 2080Ti GPU with a batch size of 8. We adopt Adam~\cite{kingma2014adam} as our optimizer, with a weight decay of $5 \times 10^{-4}$. The initial learning rate is set to $1 \times 10^{-4}$, which is reduced by a factor of 0.5 after every 5 epochs. We train both networks for 50 epochs and select the model with the best performance on the validation set as the final model. We empirically set $\alpha$, $\beta$, the IoU threshold for skipping the TPS interpolation, and $\tau$ to 0.005, 3, 0.96, and 60, respectively. 

\subsection{Ablation studies}
\begin{table}[h]
  \centering
  \caption{Ablation studies of mask prediction on the dataset proposed in~\cite{javed2017real}. ``DA'' denotes data augmentation. ``$\uparrow$'' indicates that the higher the better.}
  \begin{tabular}{c c}
    \toprule
    Method & IoU $\uparrow$\\
    \midrule
    Baseline & 0.8678\\
    Baseline + DA & 0.9408\\
    MRM + DA (ours) & \textbf{0.9520}\\
    \bottomrule
  \end{tabular}
  \label{tab:ablation_mask}
  \vspace{-3mm}
\end{table}

\begin{table}[h]
  \centering
  \caption{Ablation studies of the proposed content-aware loss. ``$\uparrow$'' indicates that the higher the better.}
    \begin{adjustbox}{width=0.37\textwidth}
    \begin{tabular}{c c c c c}
    \toprule
    $\beta$ & 0 & 3 & 6 & 9\\
    \midrule
    SSIM $\uparrow$ & 0.7212 & \textbf{0.7252} & 0.7224 & 0.7202 \\
    \bottomrule
    \end{tabular}
    \end{adjustbox}
    \label{tab:ablation_loss}
\end{table}

We consider vanilla DeepLabv3+~\cite{chen2018encoder} without data augmentation as the baseline and present the improvement we obtained in Table~\ref{tab:ablation_mask}. We validate models on the dataset proposed in~\cite{javed2017real}, which consists of 120 real-word document images. This dataset is constructed for document localization and only annotated with four corners of the document, which we use to generate the quadrilateral ground-truth mask (these document images only contain perspective deformation). As shown in Table~\ref{tab:ablation_mask}, data augmentation greatly improves the performance. The mask prediction network in our MRM also obtains improvement. The effectiveness of the introduction of the prior knowledge can be seen in Fig.~\ref{fig:MRM} (b).

We further evaluate the effectiveness of our proposed content-aware loss on our Doc3D validation set. 
We use structural similarity index (SSIM) to evaluate the quality of the rectified image that resulted from $\widehat{D}$. As shown in Table~\ref{tab:ablation_loss}, we achieve the best image quality with the setting of $\beta=3$, which indicates the contribution of our proposed content-aware loss.

\begin{table*}[t]
  \centering
  \caption{Quantitative comparisons on the DocUNet benchmark dataset~\cite{ma2018docunet}. ``\dag'' indicates that results are from publicly available images or models, the sources of which are shown in the footnotes. ``$\uparrow$'' and ``$\downarrow$'' indicate that the higher the better and the lower the better, respectively. ``$\backslash$'' indicates that results or models are not publicly available. Bold and underline denote the \textbf{best} and the \underline{second-best} results, respectively.}
  \begin{adjustbox}{width=0.79\textwidth}
  \begin{tabular}{c c c c c c c c}
  \toprule
  \multicolumn{2}{c}{\multirow{2}{*}{Method}} & \multicolumn{3}{c}{Crop}          & \multicolumn{3}{c}{Origin}                            \\ \cmidrule(lr){3-5} \cmidrule(lr){6-8}
  \multicolumn{2}{c}{}                        & MS-SSIM $\uparrow$ & LD $\downarrow$    & CER (\%)$\downarrow$          & MS-SSIM$\uparrow$          & LD $\downarrow$       & CER (\%)$\downarrow$              \\ 
  \midrule
  \multicolumn{2}{c}{DocUNet~\cite{ma2018docunet}}                  & 0.41    & 14.08 & \textbackslash{} & \textbackslash{} & \textbackslash{} & \textbackslash{} \\
  \multicolumn{2}{c}{AGUN~\cite{augn}}                    & 0.4491  & 12.08 & \textbackslash{} & \textbackslash{} & \textbackslash{} & \textbackslash{} \\ 
  \multicolumn{2}{c}{RectiNet\footnotemark[2]\dag~\cite{bandyopadhyay2021gated}}                & 0.415   & 13.2  & 35.95$\pm$20.4    & 0.3847           & 14.34            & 46.39$\pm$24.3    \\ 
  \multicolumn{2}{c}{DewarpNet\footnotemark[3]\dag~\cite{das2019dewarpnet}}               & 0.4693  & 8.98  & 21.68$\pm$20.0    & 0.4341           & 10.22            & 30.44$\pm$22.4    \\ 
  \multicolumn{2}{c}{Xie \textit{et al.}\footnotemark[4]\dag~\cite{xie2020dewarping}}              & 0.4361  & 8.50  & 76.35$\pm$31.7    & 0.1868           & 20.75            & 81.78$\pm$29.4  \\ 
  \multicolumn{2}{c}{DocProj\footnotemark[5]\dag~\cite{li2019document}}                 & 0.4071  & 11.46 & 35.95$\pm$27.0    & 0.2948           & 23.13      &  47.15$\pm$26.5  \\ 
  \multicolumn{2}{c}{PiecewiseUnwarp~\cite{das2021end}}         & \underline{0.4879}  & 9.23  & 30.01$\pm$14 & \textbackslash{} & \textbackslash{} & \textbackslash{}\\

  \multicolumn{2}{c}{DocTr\footnotemark[6]\dag~\cite{feng2021doctr}}         & \textbf{0.5085}  & 8.38  & \textbf{18.05$\pm$18.4} & \textbackslash{}
 & \textbackslash{} & \textbackslash{} \\ 
  \midrule
  
  \multicolumn{2}{c}{Marior w/o ICRM}            & 0.4380  & 9.56  & 25.88$\pm$19.9    & 0.4178           & 10.18            & 29.42$\pm$22.1    \\ 
  \multicolumn{2}{c}{Marior w/o iteraion}     & 0.4659  & \underline{8.15}  & 20.86$\pm$19.4    & \underline{0.4391}          & \underline{9.09}           & \underline{25.21$\pm$20.0}    \\ 
  \multicolumn{2}{c}{Marior}     & 0.4733  & \textbf{8.08}  & \underline{18.35$\pm$17.86}    & \textbf{0.4458} & \textbf{9.00} & \textbf{22.34$\pm$19.7}    \\ 

  \bottomrule
\end{tabular}
\end{adjustbox}
\label{tab:docunet}
\end{table*}

\subsection{Comparison on public benchmarks}
\textbf{Evaluation metrics}. We use multi-scale structural similarity (MS-SSIM)~\cite{msssim2003} and local distortion (LD)~\cite{ld2017} to evaluate the image similarity between the produced rectified image and its scanned ground truth. MS-SSIM is a widely used evaluation metric for image structural similarity. LD evaluates local distortion by predicting dense SIFT flow~\cite{liu2010sift}. We use the parameter settings of these two metrics recommended in~\cite{ma2018docunet}, which are also adopted by other methods~\cite{augn,ma2018docunet,xie2020dewarping,li2019document,das2019dewarpnet,das2021end,bandyopadhyay2021gated}. Besides, we apply Tesseract 4.1.0\footnotemark[1] with the LSTM engine as our text recognizer to recognize the text on the rectified images, which also reveals the rectification performance. We evaluate recognition results using character error rate (CER), which derives from the Levenshtein distance~\cite{levenshtein1966binary} between the recognized and reference text. The CER can be computed as $\mathrm{CER}=(s+i+d) / N$, where $s$, $i$, and $d$ are the number of substitutions, insertions, and deletions from the Levenshtein distance, respectively. $N$ is the number of characters in the reference text.

\textbf{DocUNet benchmark~\cite{ma2018docunet}}. The quantitative results on this dataset are given in Table~\ref{tab:docunet}, where “Crop” represents accurately cropped images that are usually used for comparison in previous studies. “Origin” represents originally captured images without cropping, thereby containing large marginal regions. For a more fair comparison, Faster R-CNN~\cite{ren2015faster} is used as a document detector attached to other methods when conducting experiments on “Origin” subset. The details of this detector are included in supplementary material. Text recognition is performed on 50 text-rich images as recommended in~\cite{das2019dewarpnet}. We consider the text recognized from the corresponding scanned ground-truth images as the reference text.

\footnotetext[2]{\scriptsize \url{https://github.com/DVLP-CMATERJU/RectiNet}}
\footnotetext[3]{\scriptsize \url{https://github.com/cvlab-stonybrook/DewarpNet}}
\footnotetext[4]{\scriptsize \url{https://github.com/gwxie/Dewarping-Document-Image-By-Displacement-Flow-Estimation}}
\footnotetext[5]{\scriptsize \url{https://github.com/xiaoyu258/DocProj}}
\footnotetext[6]{\scriptsize \url{https://github.com/fh2019ustc/DocTr}}

We first evaluate the effectiveness of the content rectification and iterative strategy. The results are shown in the last three rows of Table~\ref{tab:docunet}. The baseline is Marior without the ICRM (\textit{i.e.}, only adopting the MRM). After we implement content rectification in one pass without iteration (\textit{i.e.}, Marior w/o iteration in Table~\ref{tab:docunet}), all three metrics improve significantly. In particular, CER reduces by 19\% and 14\% on the ``Crop'' and ``Origin'' subsets, respectively. This demonstrates the effectiveness of the ICRM for document content rectification. Moreover, after we iteratively implement document content rectification ( \textit{i.e.}, Marior in Table~\ref{tab:docunet}), the results further improve.

\begin{figure*}[t]
  \centering
  \includegraphics[scale=0.153]{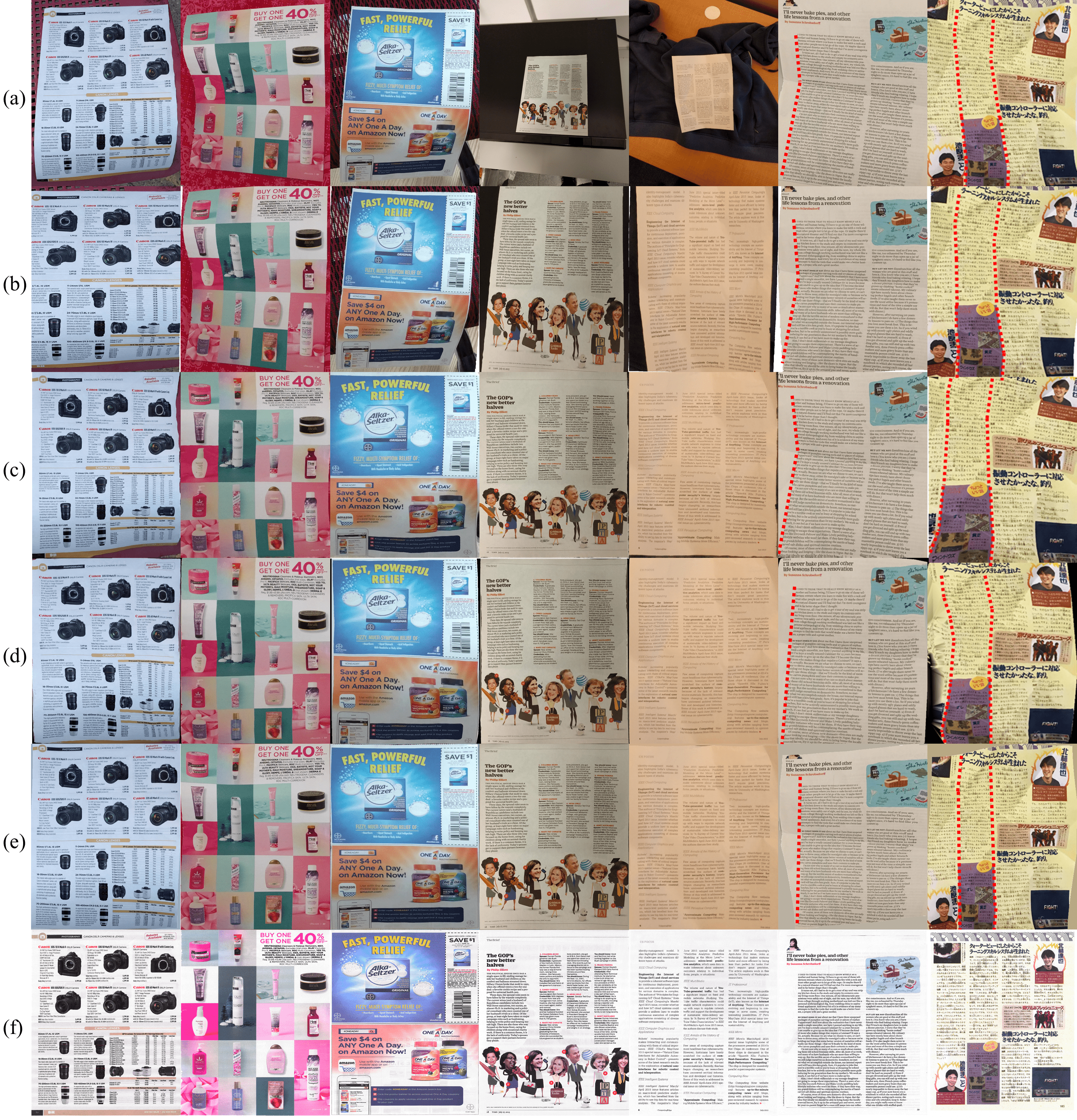}
  \caption{Qualitative comparisons on the DocUNet benchmark dataset~\cite{ma2018docunet}. (a) Input, (b) DocProj~\cite{li2019document}, (c) DewarpNet~\cite{das2019dewarpnet}, (d) method of Xie \textit{et al.}~\cite{xie2020dewarping}, (e) Marior (\textbf{ours}), and (f) scanned ground truth. Input images in the first three columns are from the ``Crop'' subset and input images in columns 4 and 5 are from the ``Origin'' subset. Input images in columns 6 and 7 are obtained by further cropping the images in the ``Crop'' subset. We highlight the deformation using red dotted lines.}
  \label{fig:compare_total}
\end{figure*}

\begin{figure*}[t]
  \centering
  \includegraphics[scale=0.135]{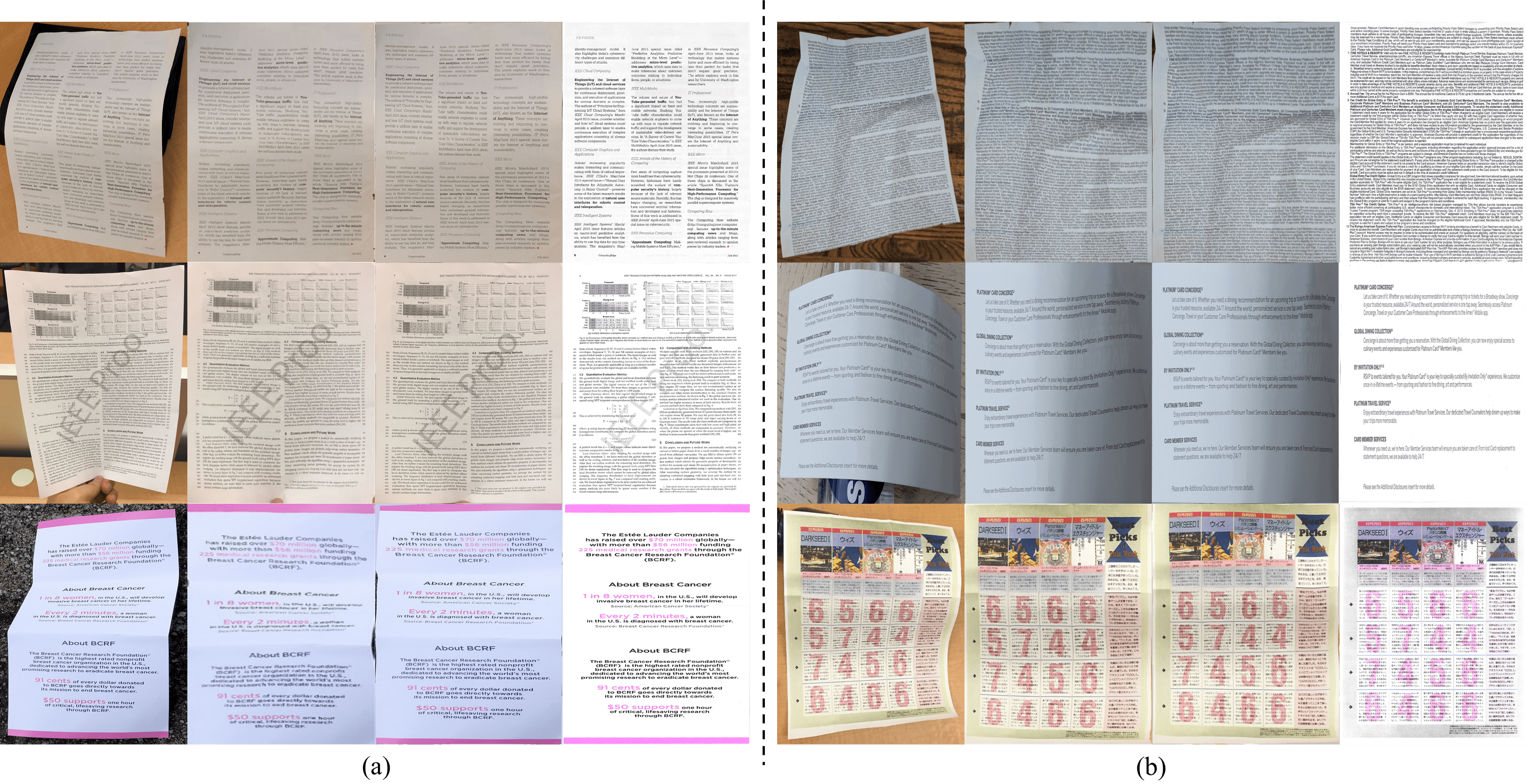}
  \caption{Qualitative comparisons with PiecewiseUnwarp~\cite{das2021end} and DocTr~\cite{feng2021doctr}. (a) From left to right are input, PiecewiseUnwarp~\cite{das2021end}, Marior (\textbf{ours}), and scanned ground truth, respectively. (b) From left to right are input, DocTr~\cite{feng2021doctr}, Marior (\textbf{ours}), and scanned ground truth, respectively.}
  \label{fig:compare_piece}
\end{figure*}

\begin{table}[t]
  \centering
  \caption{OCR evaluation results after rectification using different methods. DL denotes the abbreviation of deep learning and Reg. denotes recognizer. Bold and underline denote the \textbf{best} and the \underline{second-best} results, respectively.}
  \begin{adjustbox}{width=0.47\textwidth}
    \begin{tabular}{c c c c}
    \toprule
    \multirow{2}{*}{Method} & \multicolumn{2}{c}{CER(\%)$\downarrow$}                        & \multirow{2}{*}{Time(s)} \\ \cmidrule(lr){2-3}
                            & \multicolumn{1}{c}{Non-DL Reg.}   & DL Reg.       &                          \\ 
                            \midrule
    Source Image            & \multicolumn{1}{c}{16.12}         & 15.29         & -                        \\ 
    \midrule
    RectiNet\footnotemark[2]\dag~\cite{bandyopadhyay2021gated}                & \multicolumn{1}{c}{27.54}         & 27.57         & 1.21                     \\ 
    DewarpNet\footnotemark[3]\dag~\cite{das2019dewarpnet}               & \multicolumn{1}{c}{7.01}          & 5.46          & 0.88                     \\
    Xie \textit{et al.}\footnotemark[4]\dag~\cite{xie2020dewarping}             & \multicolumn{1}{c}{65.60}         & 11.35         & 2.01                     \\
    DocProj\footnotemark[5]\dag~\cite{li2019document}                 & \multicolumn{1}{c}{{6.13}}    & \textbf{3.12} & 6.58                     \\ 
    DocTr\footnotemark[6]\dag~\cite{feng2021doctr}                 & \multicolumn{1}{c}{{\underline{4.31}}}    & 3.69 & 2.37\\
    \midrule
    Marior w/o ICRM         & \multicolumn{1}{c}{12.26}         & 11.54         & \textbf{0.17}                     \\
    Marior w/o iteration    & \multicolumn{1}{c}{7.92}          & 7.19        & \underline{0.30}                     \\
    Marior                  & \multicolumn{1}{c}{\textbf{4.18}} & {\underline{3.28}}    & 0.85                     \\ 
    \bottomrule
    \end{tabular}
  \label{tab:ocr_real}
  \end{adjustbox}
\end{table}

When compared to existing methods on the ``Crop'' subset, Marior achieves comparable performance.However, on the ``Origin'' subset, our method outperform existing methods by a large margin even Marior is without the help of the detector. The qualitative comparisons are shown in Fig.~\ref{fig:compare_total} and~\ref{fig:compare_piece}. In Fig.~\ref{fig:compare_total}, we compare our method with DocProj~\cite{li2019document}, DewarpNet~\cite{das2019dewarpnet}, and the method of Xie \textit{et al.}~\cite{xie2020dewarping}. The input images in the first three columns are from the ``Crop'' subset. Although DocProj~\cite{li2019document} rectifies the document content to some extent, the margin remains, which results in poor visual aesthetics. DewarpNet~\cite{das2019dewarpnet} and method of Xie \textit{et al.}~\cite{xie2020dewarping} rectify the document content well and simultaneously remove the margin. Our method also achieves good perceptual performance and performs better for details than the methods in~\cite{das2019dewarpnet} and~\cite{xie2020dewarping}. The input images in the 4th and 5th columns are from the ``Origin'' subset, for which previous methods can achieve plausible results when with the help of a powerful document detector. By comparison, Marior can handle this subset withouth detector. As for input images without marginal regions in the 6th and 7th columns, Marior still achieves satisfactory performance, whereas existing methods do not.  We make further comparisons with state-of-the-art methods PiecewiseUnwarp~\cite{das2021end} and DocTr~\cite{feng2021doctr} in Fig.~\ref{fig:compare_piece}, which also demonstrates the superiority of our Marior.

\textbf{OCR\_REAL dataset~\cite{bandyopadhyay2021gated}}. This dataset contains text ground truth, which we consider as the reference text for CER metric. 
\begin{figure}[!t]
  \centering
  \includegraphics[scale=0.128]{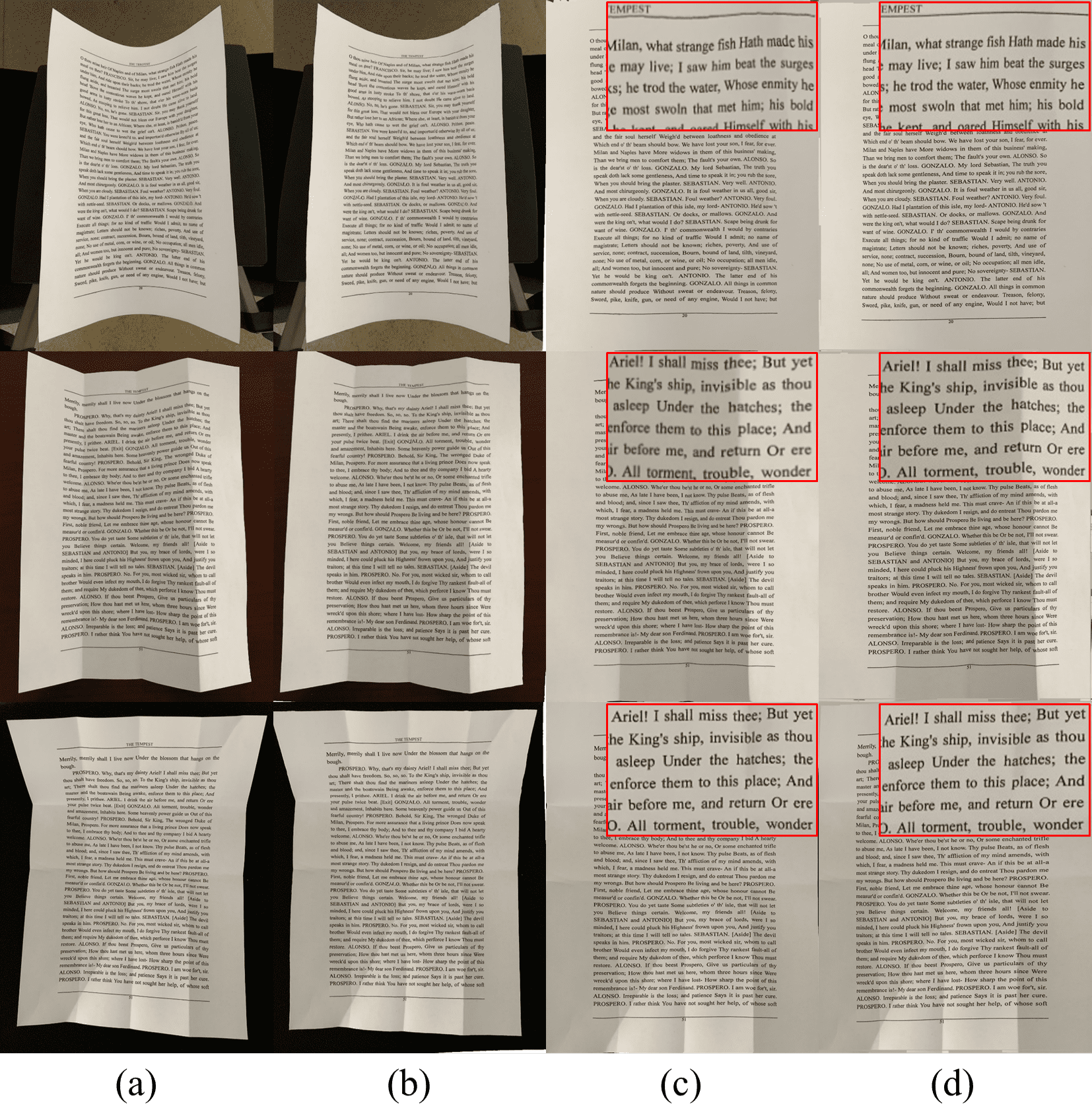}{}
  \caption{Qualitative comparisons on OCR\_REAL~\cite{bandyopadhyay2021gated}. (a) Input, (b) DocProj~\cite{li2019document}, (c) DocTr~\cite{feng2021doctr}, and (d) Marior (ours).}
  \label{fig:ocrreal}
\end{figure}
Besides, because of the lack of scanned ground-truth images, we do not evaluate the MS-SSIM and LD. Recognition performance is highly relative to the recognition engine. Therefore, to be more rigorous, we use both deep learning-based (LSTM) and non-deep learning-based engines in Tesseract 4.1.0\footnotemark[1] to perform recognition. We also evaluate the average running time of different methods on this dataset. For a fair comparison, we keep the resolution of the output image the same ($1024 \times 960$) for each method when we evaluate the running time,  which will differ when sampled images are with different resolution. The results are represented in Table~\ref{tab:ocr_real}, which shows that DocProj~\cite{li2019document}, DocTr~\cite{feng2021doctr} and Marior achieve stable and superior performance under both recognition engines when compared to the rest methods. However, DocProj~\cite{li2019document} and DocTr~\cite{feng2021doctr} is more time consuming than Marior. Additionally, as analyzed previously and shown in Fig.~\ref{fig:ocrreal},  DocProj~\cite{li2019document} fails to achieve visual aesthetics as Marior because of it's disability to remove the margin.

\section{Conclusions}
We propose a simple yet effective method, Marior, to dewarp document images in a coarse-to-fine manner. We adopt two cascaded modules to first remove the margin of the document image and then further rectify the content. The proposed Marior adaptively determines the number of iterations, thus achieving a trade-off between efficiency and performance. Our proposed method not only achieves state-of-the-art performance on the DocUNet~\cite{ma2018docunet} and OCR\_REAL~\cite{augn} benchmark datasets but also successfully tackles cases with large marginal regions and cases without marginal regions, which are less investigated in previous studies. This is a significant success in terms of dewarping documents in the wild. In future work, it is worth exploring the end-to-end optimization of the two proposed modules to achieve better performance.

\begin{acks}
This research is supported in part by NSFC (Grant No.: 61936003), GD-NSF (no.2017A030312006, No.2021A1515011870), and the Science and Technology Foundation of Guangzhou Huangpu Development District (Grant 2020GH17)
\end{acks}

\bibliographystyle{ACM-Reference-Format}
\bibliography{sigconf}

\clearpage
\noindent\textbf{\Huge Appendix}
\appendix
\section{Training DewarpNet by using data with diverse marginal situation}

\begin{figure}[h]
  \setcounter{figure}{8}
  \centering
  \includegraphics[scale=0.28]{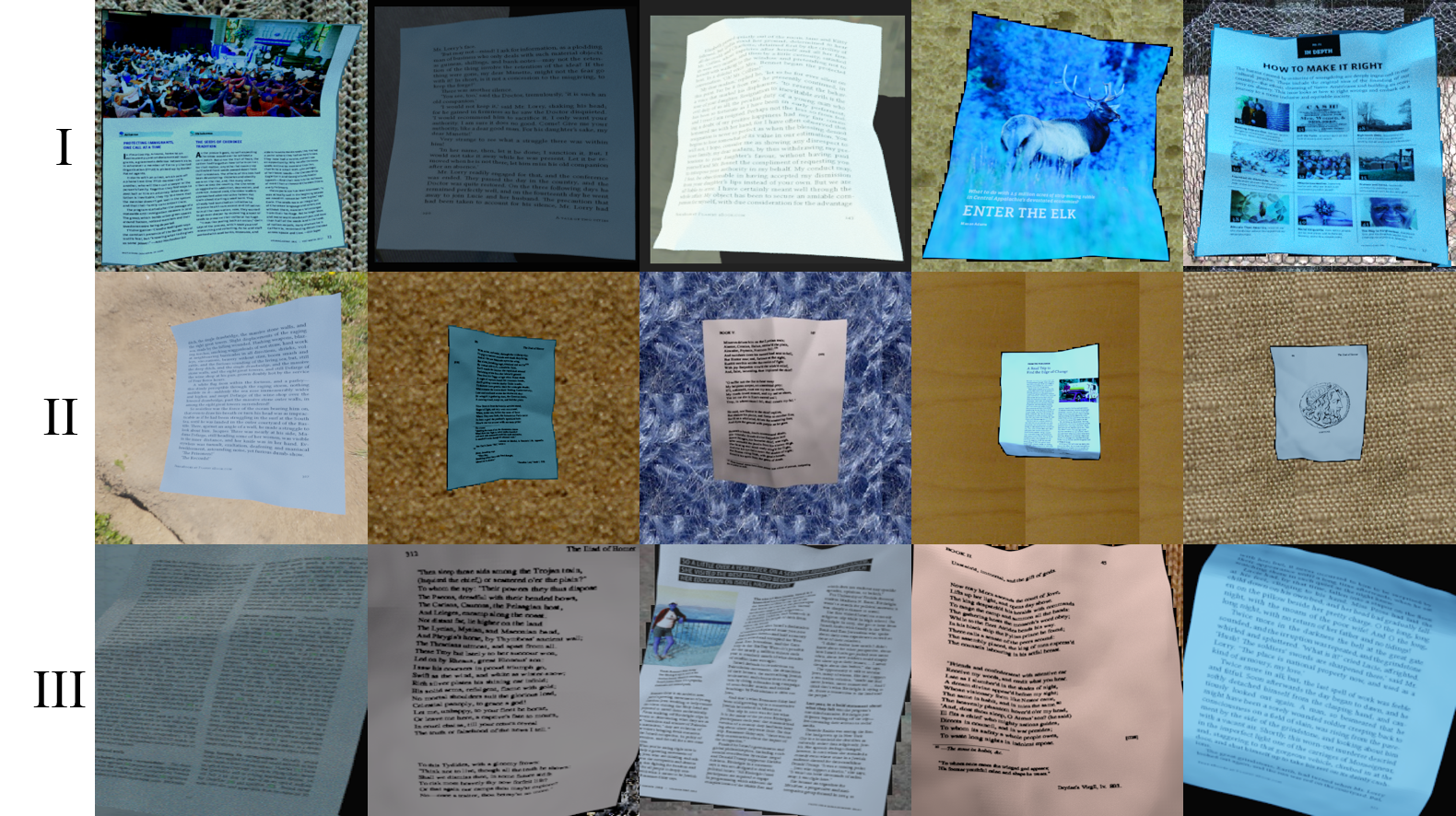}
  \caption{Traing data with different marginal situations. (I) Tightly cropped document images that are used by DewarpNet. (II) Document images with large marginal region and (III) without margins that are ignored in DewarpNet.}
  \label{fig:trainingdata}
\end{figure}

\begin{table}[!h]
\setcounter{table}{4}
\caption{Quantitative results when trainging DewarpNet with different types of data. (I), (II) and (III) represent tightly cropped images, images with large marginal regions and images without margins as in Fig.~\ref{fig:trainingdata}, respectively. }
\begin{tabular}{lllllll}
\hline
\multicolumn{3}{c}{Training data} & \multicolumn{2}{c}{Crop} & \multicolumn{2}{c}{Origin} \\ \cmidrule(lr){1-3} \cmidrule(lr){4-5} \cmidrule(lr){6-7}
\multicolumn{1}{c}{I} & \multicolumn{1}{c}{II} & \multicolumn{1}{c}{III} & \multicolumn{1}{c}{MS-SSIM↑} & \multicolumn{1}{c}{LD↓} & \multicolumn{1}{c}{MS-SSIM↑} & \multicolumn{1}{c}{LD↓} \\ \hline
\multicolumn{1}{c}{\checkmark} & \multicolumn{1}{c}{-} & \multicolumn{1}{c}{-} & \multicolumn{1}{c}{0.4637} & \multicolumn{1}{c}{9.20} & \multicolumn{1}{c}{0.2055} & \multicolumn{1}{c}{56.87} \\
\multicolumn{1}{c}{\checkmark} & \multicolumn{1}{c}{\checkmark} & \multicolumn{1}{c}{-} & \multicolumn{1}{c}{0.4666} & \multicolumn{1}{c}{9.01} & \multicolumn{1}{c}{0.3842} & \multicolumn{1}{c}{17.98} \\
\multicolumn{1}{c}{\checkmark} & \multicolumn{1}{c}{-} & \multicolumn{1}{c}{\checkmark} & \multicolumn{1}{c}{0.4405} & \multicolumn{1}{c}{10.56} & \multicolumn{1}{c}{0.1631} & \multicolumn{1}{c}{45.84} \\
\multicolumn{1}{c}{\checkmark} & \multicolumn{1}{c}{\checkmark} & \multicolumn{1}{c}{\checkmark} & \multicolumn{1}{c}{0.4372} & \multicolumn{1}{c}{10.41} & \multicolumn{1}{c}{0.3644} & \multicolumn{1}{c}{20.18} \\ \hline
\end{tabular}
\label{tab:dewarpnet}
\end{table}

To enable models to handle images with diverse marginal situations, the most intuitive way is to take all these situations into consideration during training. As only considering dewarping tightly cropped document images, DewarpNet~\cite{das2019dewarpnet} only takes images like Fig.~\ref{fig:trainingdata} (I) as training data while ignoring images with large marginal region or whithout margins like Fig.~\ref{fig:trainingdata} (II) and (III). Here we include data in Fig.~\ref{fig:trainingdata} (II) and (III) to retrain DewarpNet~\cite{das2019dewarpnet}. Note that all experiments are based on the public available training code of DewarpNet~\cite{das2019dewarpnet} and the only difference between these experiments is the type of training data. Results are given in Table~\ref{tab:dewarpnet}, from which we can see that taking all marginal situations into consideration during training does not result in satisfactory performance. This is mainly due to the difficulty in predicting 3D coordinates and backward mapping of document when the depth and scale of document have large variations. 
The model implicitly learns to identify the foreground document and remove marginal region. In contrast, segmentation is a more easy and straightforward task when countered with these variations, which is adopted in Marior. Marior decouple the margin removal process and make the displacement flow prediction network focused on content rectification. 

\section{Architecture of the displacement flow prediction network}
\begin{figure}[h]
  \centering
  \includegraphics[scale=0.27]{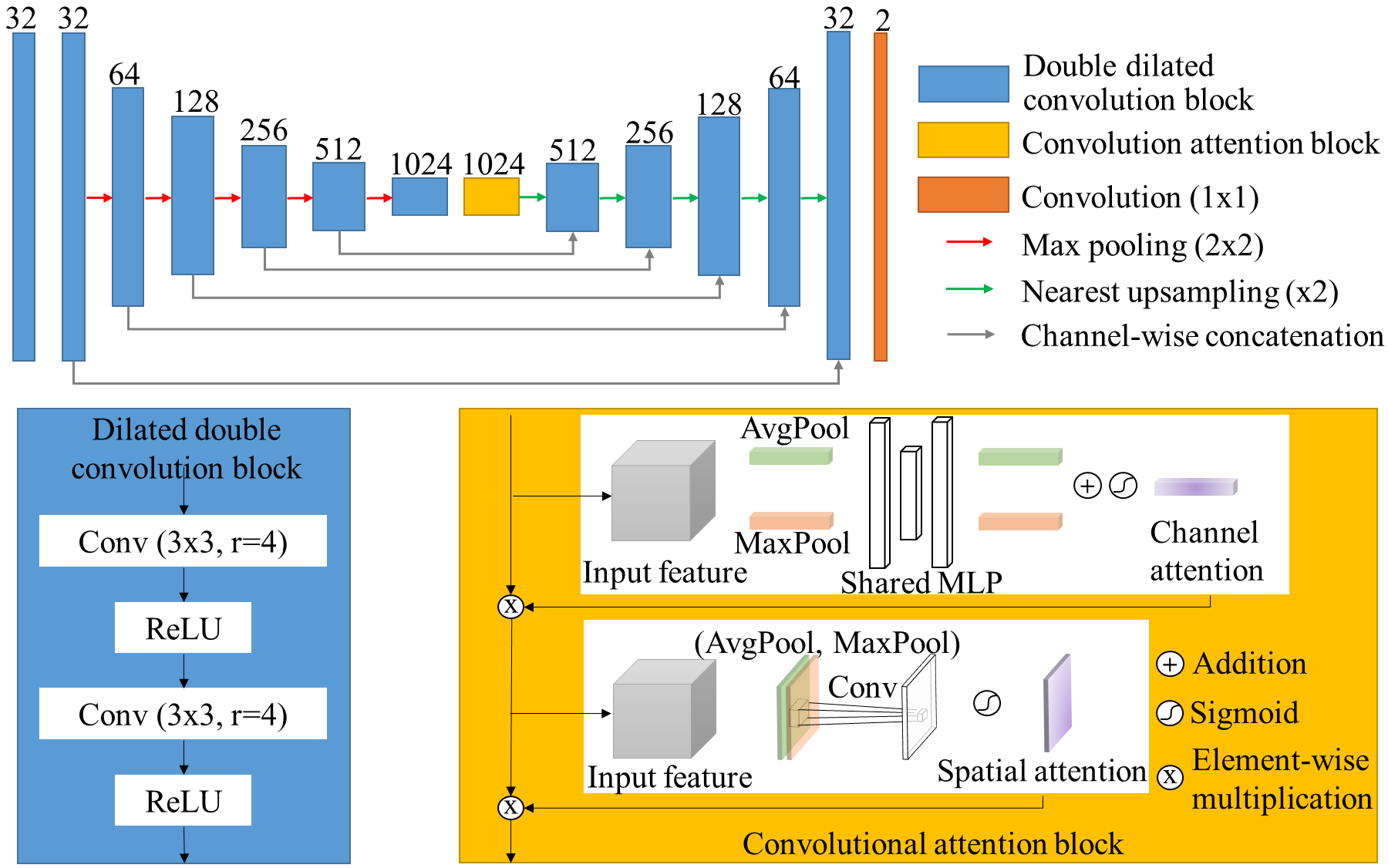}
  \caption{The architecture of our displacement flow prediction network.}
  \label{fig:architecture}
\end{figure}

We adopt the commonly used encoder-decoder with skip connections as our displacement flow prediction network. We implement the attention strategy~\cite{woo2018cbam} in bottleneck and dilated convolution~\cite{yu2015multi} to enlarge receptive filed and capture global information. The detail architecture is shown in Fig.~\ref{fig:architecture}.

\section{The effect of different iterations }
\begin{figure}[h]
  \centering
  \includegraphics[scale=0.38]{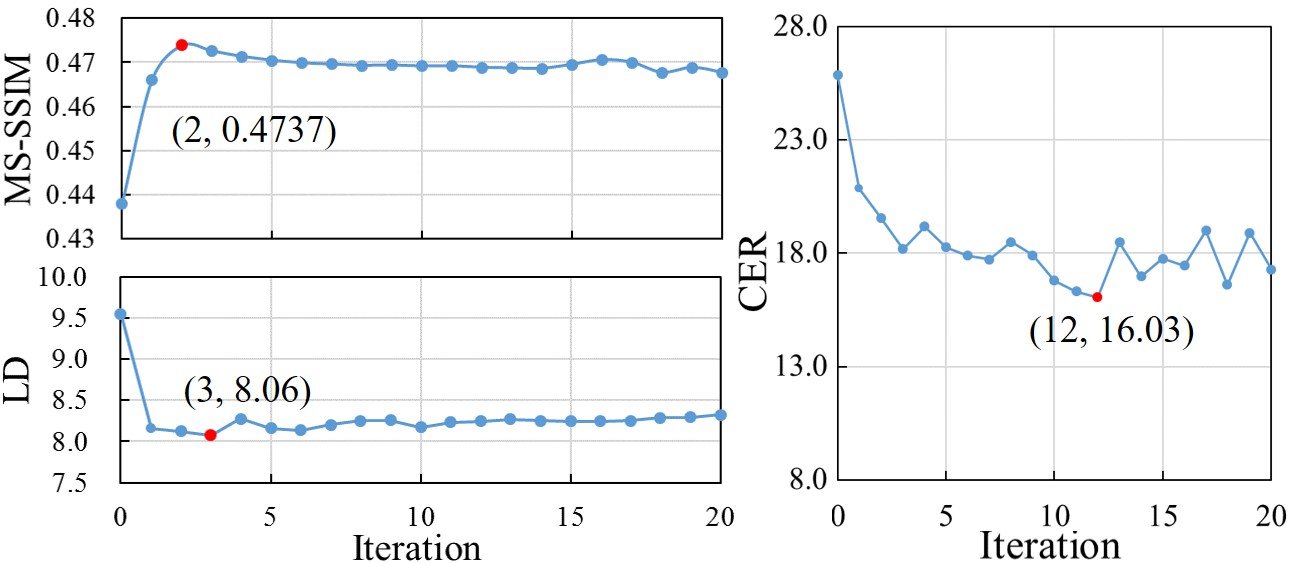}
  \caption{Dewarping performance under different iterations. We mark the best with red dot.}
  \label{fig:iteration_metric}
\end{figure}

\begin{figure}[t]
  \centering
  \includegraphics[scale=0.39]{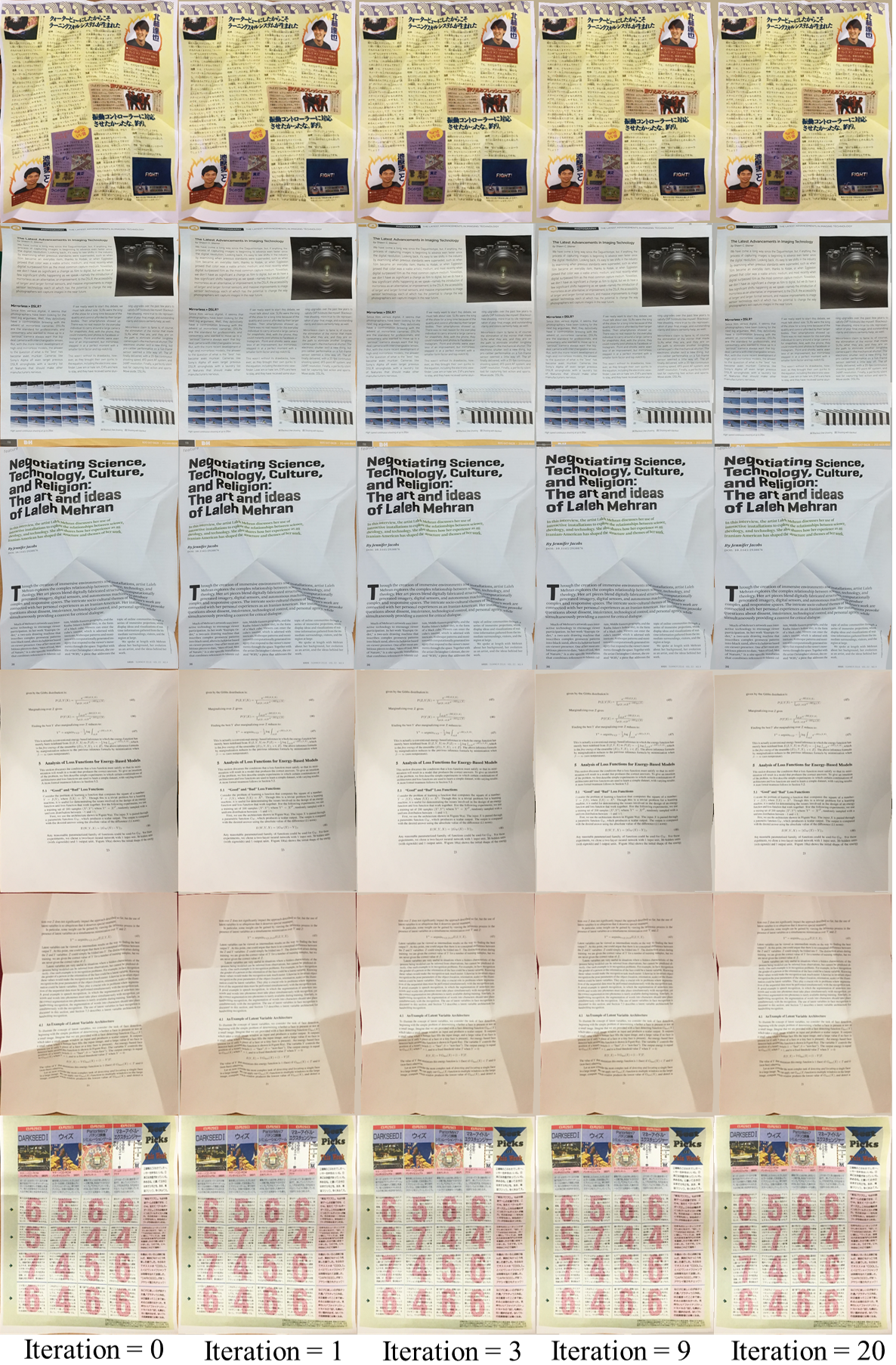}
  \caption{Qualitative results under different iterations.}
  \label{fig:iteration_visualize}
\end{figure}

We explore the effect of different iterations when we exclude our adaptive termination method. Results shown in Fig.~\ref{fig:iteration_metric} demonstrate that our iterative strategy can progressively improve metrics in the first few iterations. However, as the iteration continues, performance of these two metrics no longer improve and even deteriorate. So we terminate iteration by using our adaptive strategy. We also give some qualitative results in Fig.~\ref{fig:iteration_visualize}.

\section{The details of implementing document detector}

We adopt Faster R-CNN~\cite{ren2015faster} as our document detector, which is trained by using Doc3D~\cite{das2019dewarpnet} dataset. The ground-truth bounding box derives from the minimum bounding rectangle of the document mask. The resulting model achieves 98.9\% average precision with IoU threshold of 75\% ($AP_{75}$) on dataset proposed in~\cite{javed2017real}. We apply this detector to crop the document out when we conduct experiments on the ``Origin'' subset. To be more consistent with the ``Crop'' subset, we enlarge the predicted bounding box by 30 pixels in all four directions. We consider the cropped image deriving from these bounding box as the input image for  dewarping models. Results are given in Table~\ref{tab:detection}. Marior without detector achieves better performance than other method with detector. When combined with the detector, Marior's performance improve further. 

\begin{table}[h]
\centering
\caption{Results on the ``Origin'' subset. ``*'' denotes that the document detector is applied before dewarping. ``\dag'' indicates that results are from publicly available images or models.}
\begin{adjustbox}{width=0.45\textwidth}
\begin{tabular}{cccc}
\hline
                     & MS-SSIM↑ & LD↓   & CER (\%)↓  \\ \hline
RectiNet\dag~\cite{bandyopadhyay2021gated}   & 0.2370   & 44.13 & 77.72±12.6 \\
RectiNet\dag*~\cite{bandyopadhyay2021gated}   & 0.3847   & 14.34 & 46.39±24.3 \\
DewarpNet\dag~\cite{das2019dewarpnet}  & 0.2022   & 53.16 & 70.98±26.0 \\
DewarpNet\dag*~\cite{das2019dewarpnet}  & 0.4341   & 10.22 & 30.44±22.4 \\
Xie et al.\dag~\cite{xie2020dewarping} & 0.1001 &42.16& 99.28±1.9 \\
Xie et al.\dag*~\cite{xie2020dewarping} & 0.1868   & 20.75 & 81.78±29.4 \\
DocProj\dag~\cite{li2019document}   & 0.1799 &52.97 &87.54±22.2 \\
DocProj\dag*~\cite{li2019document}   & 0.2948   & 23.13 & 47.15±26.5 \\ \hline
Marior  & \underline{0.4458} & \underline{9.00} &\underline{22.34±19.7} \\
Marior* &  \textbf{0.4666}   &  \textbf{8.38}  & \textbf{20.22±19.6} \\\hline
\end{tabular}
\end{adjustbox}
\label{tab:detection}
\end{table}

\section{Performance of previous methods on images pre-dewarped by MRM}
MRM can also be a pre-prossing module for other methods. We give results of previous studies on images pre-dewarped by MRM as below, which demonstrate the superiority of ICRM over previous methods.

\begin{table}[h]
\centering
\caption{Performance of previous methods on images pre-dewarped by MRM.}
\begin{adjustbox}{width=0.45\textwidth}
\begin{tabular}{ccccc}
\hline
 & \multicolumn{2}{c}{Crop} & \multicolumn{2}{c}{Origin} \\ \cmidrule(lr){2-3} \cmidrule(lr){4-5}
 & MS-SSIM & LD & MS-SSIM & LD \\ \hline
MRM+RectiNet & 0.4061 & 12.37 & 0.4024 & 12.46 \\
MRM+DewarpNet & 0.4211 & 10.90 & 0.4186 & 10.95 \\
MRM+Xie et al. & 0.1673 & 22.07 & 0.1824 & 21.71 \\
MRM+DocProj & 0.2725 & 13.71 & 0.2785 & 13.66 \\ \hline
MRM+ICRM (Marior) & \textbf{0.4733} & \textbf{8.08} & \textbf{0.4458} & \textbf{9.00} \\ \hline
\end{tabular}
\end{adjustbox}
\end{table}

\end{document}